\documentclass[letterpaper, 10 pt, conference]{ieeeconf}  

\IEEEoverridecommandlockouts                              

\usepackage{amsmath,amsfonts,amsthm,graphicx,float,hyperref,algorithm}
\usepackage{bbm}
\usepackage{booktabs}
\usepackage{multirow}
\usepackage{threeparttable}
\usepackage{longtable}

\usepackage{algorithmic}

\usepackage[utf8]{inputenc}
\usepackage[english]{babel}

\usepackage{xcolor}

\title{\LARGE \bf
 Early ICU Mortality Prediction and Survival Analysis for Respiratory Failure
}

\author{Yilin Yin and Chun-An Chou
\thanks{The authors are with the Department of Mechanical and Industrial Engineering, Northeastern University, Boston, MA 02115 USA (e-mail: yin.yil@northeastern.edu; ch.chou@northeastern.edu).}
}

\begin{document}

\maketitle
\thispagestyle{empty}
\pagestyle{empty}

\begin{abstract}
 Respiratory failure is the one of major causes of death in critical care unit. During the outbreak of COVID-19, critical care units experienced an extreme shortage of mechanical ventilation because of respiratory failure related syndromes. To help this, the early mortality risk prediction in patients who suffer respiratory failure can provide timely support for clinical treatment and resource management. In the study, we propose a dynamic modeling approach for early mortality risk prediction of the respiratory failure patients based on the first 24 hours ICU physiological data. Our proposed model is validated on the eICU collaborate database. We achieved a high AUROC performance (80-83\%) and significantly improved AUCPR 4\% on Day 5 since ICU admission, compared to the state-of-art prediction models. In addition, we illustrated that the survival curve includes the time-varying information for the early ICU admission survival analysis.    



\end{abstract}

\section{INTRODUCTION}
    Mortality risk prediction is one of the essential issues of healthcare decision making. The acute respiratory distress syndrome (ARDS) is caused by respiratory failure, which has 55\% mortality rate for general ICU admissions \cite{bellani2016epidemiology}. During the pandemic of coronavirus disease (COVID-19), one third of the COVID-19 patients suffered the respiratory  failure \cite{bolourani2021machine}. The respiratory failure is one of most severe causes of death with 39\% average mortality rate among the COVID-19 patients with ARDS \cite{hasan2020mortality}. In this situation, overwhelmed intensive care unit (ICU) system requires an urgent need of accurate mortality risk prediction for respiratory failure patients in order to allocate clinical resources during ICU stays and reduce morbidity accordingly. \par

On the other hand, the existing literature is limited on the research of ICU mortality prediction related to respiratory failure patients. In previous studies, quantitative tools have been used for estimating the mortality risks using data collected from respiratory failure patients in ICU, including two major directions: first, the traditional methods of mortality prediction are scoring systems, including Simpler Acute Physiology Score (SAPS) \cite{saps1}, Acute Physiology and Chronic Health Evaluation (APACHE) \cite{Knaus1991TheAI}, and Sequential Organ Failure Assessment (SOFA) \cite{sofa_2001}. In study \cite{aydogdu2010mortality,saleh2015comparison}, these risk scoring systems have been compared among the ICU patients with ARDS. The overall performance was limited, given that these scoring systems were developed by empirical models for general ICU patients population. These scoring system were not specific designed for respiratory failure patients. Thus, some studies have developed data-driven models using machine learning techniques \cite{gannon2018outcomes,bolourani2021machine}, which could be flexibly applied to various types of critical illness including respiratory failure. However, the conventional machine learning-based predictive methods have less consideration on the time-varying information while the patient's risk probability is highly time-dependent. \par

To takle the above-mentioned challenges, we propose a cumulative hazard function based autoregressive hidden Markov model (CHF-AR-HMM) to handle the time-varying mortality risks. Additionally, this model has the advantages of estimating the short term risk against data imbalance and sparsity. Different from our previous work \cite{yin2021novel}, the current model learns the survival model parameter cumulative hazard function in each time windows instead of directly estimating the parameter based on distribution of length of survival. Thus the cumulative hazard function has an increased capability to reflect the time-variation. \par

This paper is organized as follows. In Section \ref{secii} the model is introduced. In Section \ref{seciii}, the  experiments and results are demonstrated. A conclusion is given in Section \ref{seciv}.

\begin{figure}[htbp]
  \centering 
  \includegraphics[width=0.45\textwidth]{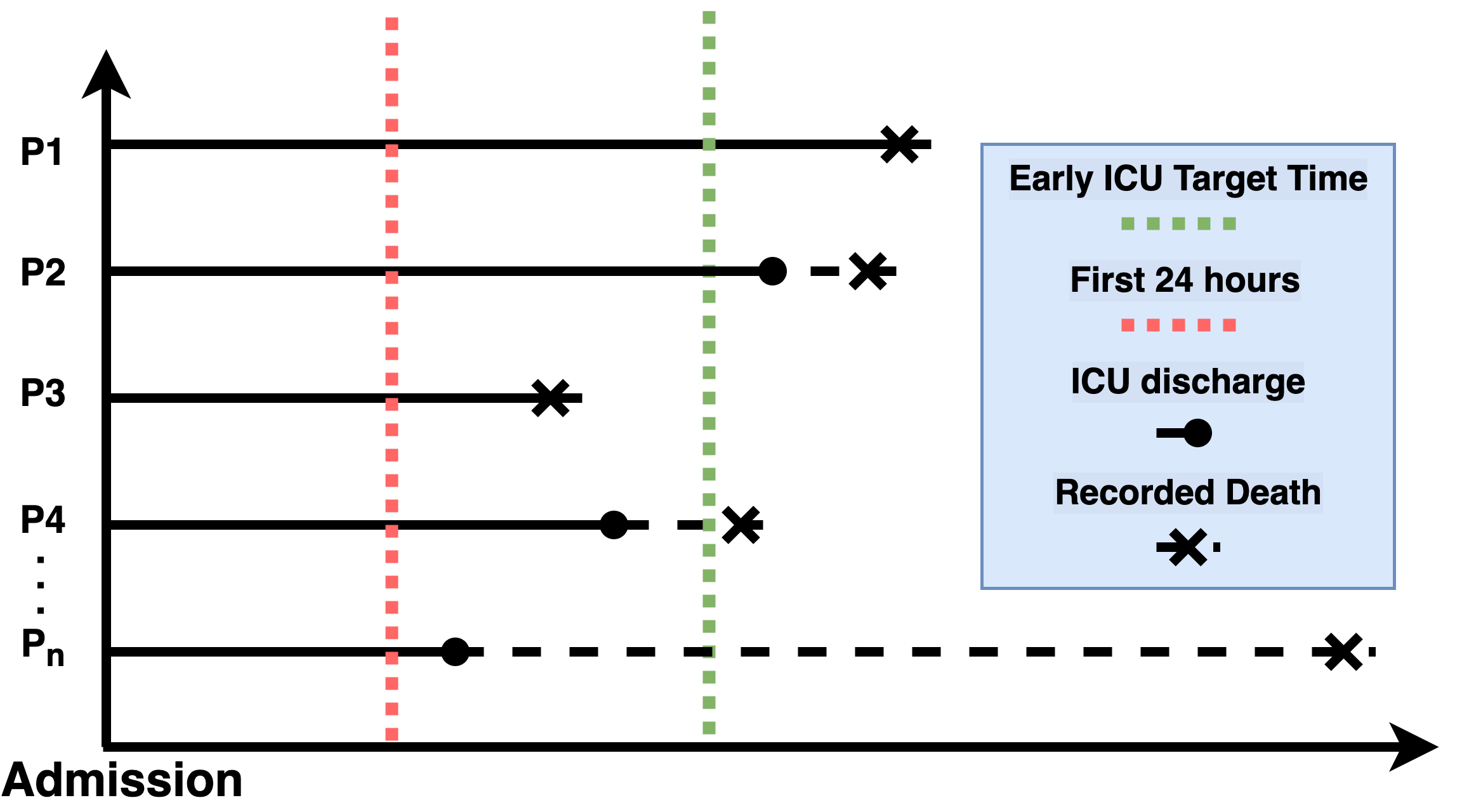} 
  \caption{The illustration of patient discharge from ICU before or survived after target time (green line).}
  \label{censor}
\end{figure}

\section{Methodology}
\label{secii}
 \begin{figure*}[htbp]
  \centering 
  \includegraphics[width=0.8\textwidth]{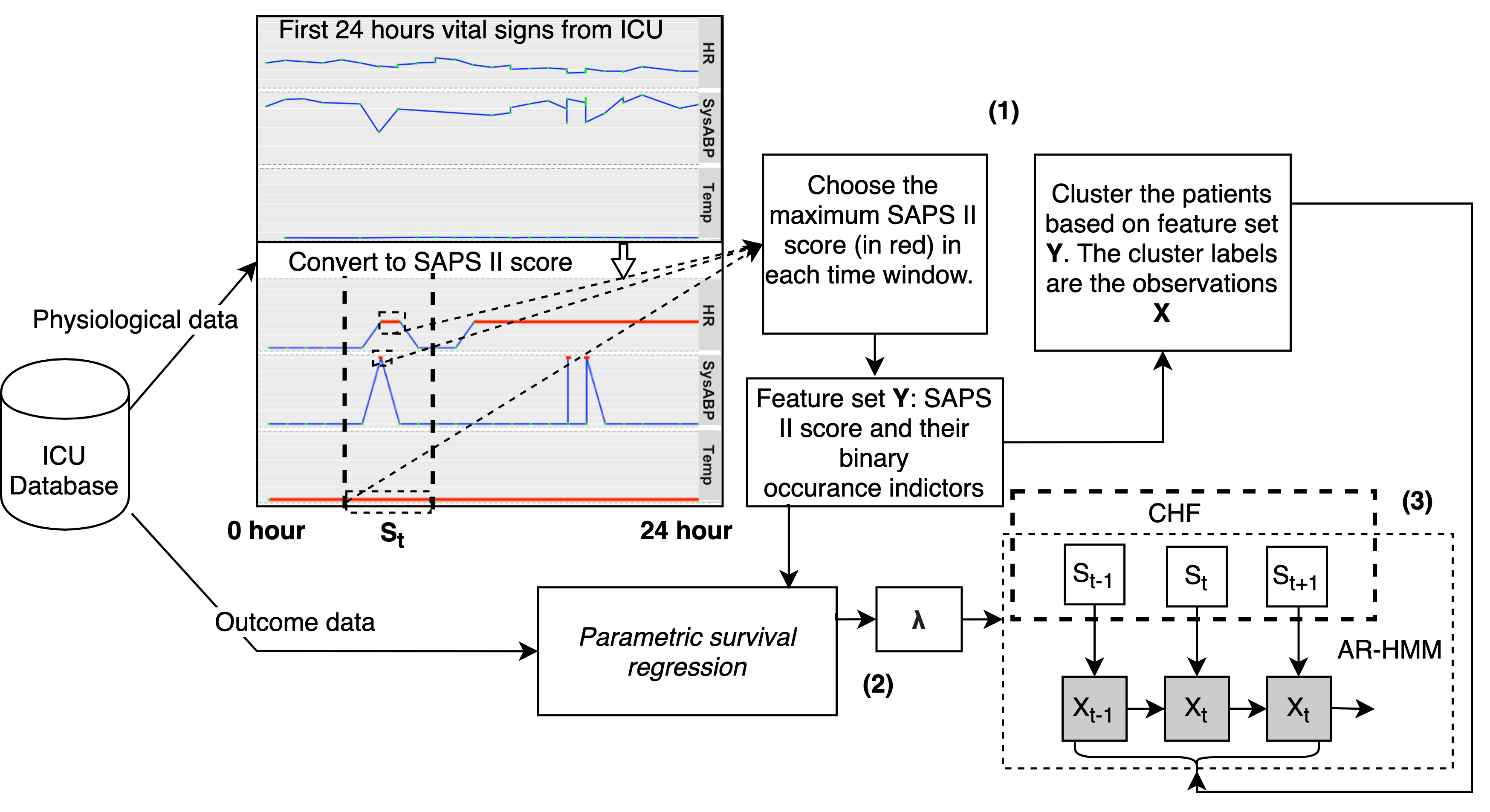} 
  \caption{The modeling framework includes three steps: (1) feature engineering; (2) transition probability estimation; and (4) dynamical model construction.}
  \label{CHFARHMM}
\end{figure*}

\subsection{Feature Engineering}
\label{SecA}

Our model uses the first 24 hours ICU physiological data to predict the mortality risk by certain target time as shown in Figure \ref{censor}. Firstly, we segment the first 24 hour data into equal sized time windows. To avoid loss of information with the low sampling rate data, the size of time window is pre-determined by hours ($n = \{1, 2,...,24\}$). In order to sort the risk for each variable, the SAPS II is employed to discretize the original physiological values into integer scores. While the higher SAPS II scores reflects a greater risk, the highest score in each time window will corresponds to the worst case representative for variables denoted as $\textbf{Y}^j_t = [y^1_t,..., y^p_T]$, where $p$ is the number of ICU variables and $T$ is the number of time windows. After data discretization, the existing missing data of each time window has been considered as a new feature set to reflect the sparsity of original dataset. The binary occurrence indicator $b^j_t \in \{0, 1\}$ marks the missing values for each variable, which defined as as 0 if missing and otherwise 1. The feature matrix for individual patients is denoted as ${\textbf{Y}^j_t} = [y^1_t,..., y^p_T,b^1_t,..., b^p_T]$. \par

Afterwards, in order to measure the variation of the observation sequence, the feature sequences from the feature matrix ${\textbf{Y}^j_t}$ are labeled. We apply the partition around medoids (PAM) \cite{pam} to cluster the feature sequences across the time windows with the highest similarity measured by Gower's distance \cite{gower1971general}. $\textbf{X} \in \{1, ..., K\}$ denoted as the number of clusters are now the integer values of the observation sequence along the time window.

\subsection{Transition Probability}
\label{SecB}
In our study, the hidden states $S \in \{Survival, Death\}$ are defined as the future survival status by the target time $D$, which is the terminal of mortality measurement. On the other hand, we lose the assumption of the AR-HMM that the current state is not directly correlated to the previous state, where the observation still follows $\textbf{X}_t = f_{S_t}(\textbf{X}_{t-1})$, where $f_{S_t}$ is the autoregressive function determined by corresponding hidden state at time $t$. $Death$ is not considered as an absorbing state. In this case, the transition probability is the prior probability of $Death$:

\begin{align}
\label{eqn:eqlabel1}
\begin{split}
\hat{\theta}_{{Survival, Death} -> Death} & = P(S_t = Death|t,\lambda_t) \\
& = 1-e^{-\lambda_t V_t},
\end{split}
\end{align}
\noindent where $V_t = D + n*t$ is the total duration by time window $t$ with window size $n$ toward the target time $D$. $\lambda_t$ is the hazard function parameter at time window $t$ learned by the exponential parametric survival regression:

\begin{align}
\label{eqn:eqlabel2}
\begin{split}
\lambda_t = e^{\beta_t \textbf{Y}_t},
\end{split}
\end{align}
\noindent where $\beta_t$ is the vector of coefficient for feature sequence at time window $t$. This model is learned by response variables including the length of survival and the outcome variable labeling the death by target time and random censoring patients (discharge from ICU before or censored by the target time).  

For the training data, the last hidden state in the sequence is labeled as same as the outcome, where the data censoring by target time is considered as survival. The rest of the hidden states $S_t$ are determined by the normalized probability of death. In the density based supervised normalization process, we first fit the probability of death from the training data by the outcome label in each time window with Gaussian density estimation. Thus, for both death and survival labels, the possible data points from probability of death of density have been mapped into the density. Then, we measure the proportion of the density of death class among the total density at each corresponding data point from original probability of death. This proportion is now the normalized version of probability of death. We can map the original probability of death to the corresponding proportion of the density as the new probability of death for both training and testing data to classify the hidden states with cut-off 0.5.

\begin{figure}[htbp!]
  \centering \vspace{0.5cm}
  \includegraphics[width=0.3\textwidth]{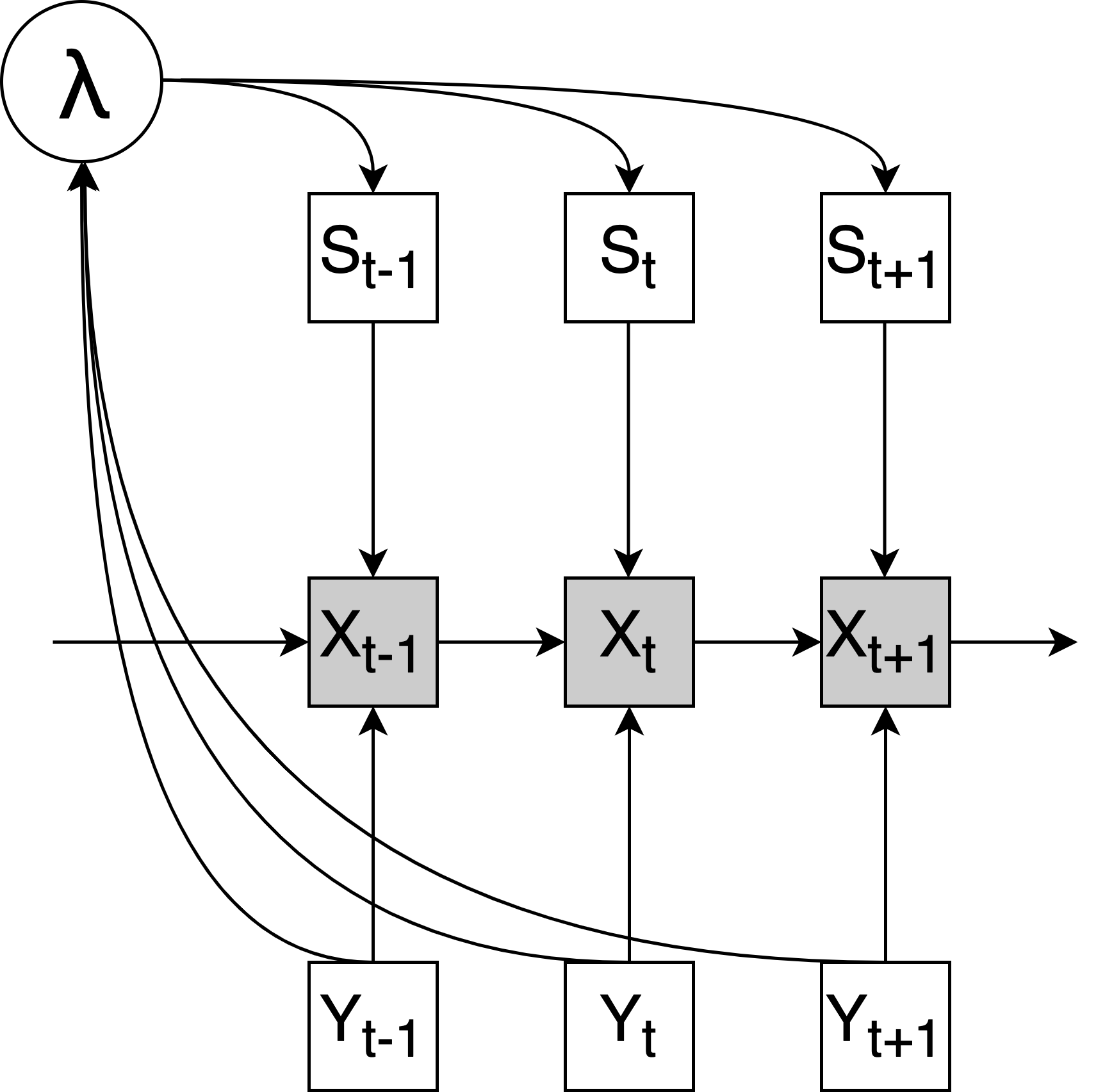} 
  \caption{CHF-AR-HMM diagram.} \vspace{-0.3cm}
  \label{CHF-AR-HMM}
\end{figure}

\subsection{CHF-AR-HMM}
\label{SecC}

The structure of CHF-AR-HMM is shown in Figure \ref{CHF-AR-HMM}. CHF-AR-HMM generate the joint probability $\Psi$ of all the hidden states and observations:

\begin{align}
\label{eqn:eqlabel9}
\begin{split}
 \Psi & = \theta_{S_1} \phi_{X_1|S_1} \prod^T_{t=2} \theta_{S_t}~\phi_{({X_t|X_{t-1},S_t})}.
\end{split}
\end{align}

\noindent where $\phi$ is emission probability. Recall the autoregressive function $f_{S_t}$, we generalize the concept to the prior probability about the correlation between observations and the hidden states. The emission probability $\phi_{l|k,S_t} = P(X_t = l|X_{t-1} = k , S_t)$ determines the prior probability of current observation $X_t$ in the condition of the current state $S_t$ with previous observation $X_{t-1}$, where $l,k \in \{1,...,K\}$. $\Psi_{Death}$ is identified as joint probability of high mortality risk group where the hidden states sequence contains $Death$. $\Psi_{Survival}$ is for low risk group with only $Survival$ hidden state. Based on the $\Psi_{Death}$ and $\Psi_{Survival}$, we define $\eta$ as the mortality risk measurement by target time:

\begin{align}
\label{eqn:eqlabel10}
\begin{split}
\eta &= {\frac{\Sigma \Psi_{Death}}{\Sigma \Psi_{Death} + \Sigma\Psi_{Survival}}},
\end{split}
\end{align}
\noindent where we sum up all the joint probabilities for either high or low risk group.


\section{Experimental Results}
\label{seciii}

Our model is validated on the eICU Collaborative Research Database \cite{pollard2018eicu} using the subset of patients who have been diagnosed with respiratory failure and have a minimum of 24 hours ICU stay. The samples with incomplete data for heart rate, blood pressure and Glasgow Coma Scale across the time window have been disregarded, and thus a total of 4391 samples are included. Each time window is fixed to 12 hours in this study. The rest missing values of each variable are imputed with their median. Because the lower bound of median stay in ICU for COVID-19 patients is 5 days \cite{rees2020covid}, the target time is set from Day 2 to Day 5 after ICU admission to cover the early risks.

\subsection{Evaluation}
\label{seciiia}
In this study we evaluate the model with three metrics. The primary metric is area under precision-recall curve (AUCPR), which is informative when positive class is more important. The baseline for AUCPR is equivalent to the ratio of death class. Secondly, since $\eta$ is equivalent to the time-to-event joint probability for survival analysis, the concordance-statistic (C-statistic) is used to measure the goodness of fit for the binary classification. The concordance is defined that the joint probability of risk should be higher while the length of survival is shorter comparing to other samples. Finally, we still evaluate the model by AUROC considering the equal importance of both class.  We include exponential parametric survival regression, logistic regression and SAPS II scoring system as baseline methods. The exponential parametric survival regression generates the hazard function to get the probability of death by target time. The maximum SAPS II score of variables in first 24 hours are used for baselines prediction. Our model is trained and evaluated by the 30 times 3-fold cross-validation on 2 to 5 days since ICU admission respectively.  

\begin{table}[htbp!]
\tiny \addtolength{\tabcolsep}{-4pt} 
\centering
\caption{Performance comparison based on 12-hour time interval.}
\label{output_table}
\begin{tabular}{ccccccccc}
\toprule
\textbf{Model}                        & \textbf{} & \textbf{} & \textbf{AUCPR} & \textbf{95\% CI} & \textbf{C-statistic} & \textbf{95\% CI} & \textbf{AUROC} & \textbf{95\% CI} \\ \midrule
\multirow{3}{*}{\textbf{CHF-AR-HMM}}     
                                      &                    &  \textbf{Day 2}              & \textbf{0.27}   & (0.26, 0.28)   & \textbf{0.82}  & (0.81, 0.83)   & \textbf{0.83}  & (0.82, 0.83)
                                                             \\
&              & \textbf{Day 3}            & \textbf{0.35}  & (0.33, 0.35) & \textbf{0.80}  & (0.80, 0.81)  & \textbf{0.81}  & (0.81, 0.82)
 \\
&                     & \textbf{Day 4}            & \textbf{0.42}  & (0.41, 0.43)  & \textbf{0.79}  & (0.79, 0.80)  & \textbf{0.81}  & (0.80, 0.82)
 \\
&                     & \textbf{Day 5}                  & \textbf{0.44}  & (0.43, 0.45)  & \textbf{0.78}  & (0.78, 0.79)  & \textbf{0.80}  &  (0.79, 0.80)
\\

                                      \midrule                                      
\multirow{4}{*}{\textbf{SAPS II}}                                             &                    & \textbf{Day 2}              & 0.22 & (0.21, 0.24)  & 0.79  & (0.78, 0.80)  & 0.80  & (0.79, 0.81) \\
&            & \textbf{Day 3}            & 0.29  & (0.27, 0.30)  & 0.76  & (0.76, 0.77)  & 0.78  &  (0.77, 0.79)
 \\
&                     & \textbf{Day 4}            & 0.35  & (0.34, 0.36)  & 0.75  & (0.74, 0.76)  & 0.77  &  (0.77, 0.78)
 \\
&                     & \textbf{Day 5}                  & 0.37  & (0.36, 0.38)  & 0.74  & (0.73, 0.74)  & 0.76  &  (0.75, 0.76)
\\
\midrule                                      
\multirow{4}{*}{\textbf{Parametric Survival Regression}}                                             &                    & \textbf{Day 2}              & 0.25  & (0.23, 0.26)  & 0.81  & (0.80, 0.81)  & 0.81  &  (0.81, 0.82) \\
&            & \textbf{Day 3}            & 0.32  & (0.30, 0.33) & 0.78  & (0.78, 0.79)  & 0.80  &  (0.79, 0.81)
 \\
&                     & \textbf{Day 4}            & 0.38  & (0.37, 0.40)  & 0.78  & (0.77, 0.78)  & 0.79  & (0.78, 0.80)
 \\
&                     & \textbf{Day 5}                  & 0.40  & (0.38, 0.41)  & 0.76  & (0.75, 0.77)  & 0.78  & (0.77, 0.78)
\\
\midrule                                      
\multirow{4}{*}{\textbf{Logistic Regression}}                                             &                    & \textbf{Day 2}              & 0.23  & (0.22, 0.25)  & 0.80  & (0.80, 0.81)  & 0.81  &  (0.80, 0.82) \\
&            & \textbf{Day 3}            & 0.31  & (0.30, 0.33)  & 0.78  & (0.78, 0.79)  & 0.80  &  (0.79, 0.80)
 \\
&                     & \textbf{Day 4}            & 0.38  & (0.36, 0.39)  & 0.77  & (0.77, 0.78)  & 0.79  & (0.78, 0.80)
 \\
&                     & \textbf{Day 5}                  & 0.39  & (0.38, 0.41)  & 0.76  & (0.75, 0.76)  & 0.77  & (0.77, 0.78)

\\ \bottomrule
\end{tabular}
\begin{tablenotes}
  \item[] The AUCPR has the baseline value from 0.05, 0.09, 0.12 and 0.15 for Day 2 to Day 5 respectively. The C-statistic and AUROC has the baseline value at 0.5.
\end{tablenotes}
\end{table}    

\subsection{Performance Comparison}
\label{seciiib}

\begin{figure}[htbp]
  \centering 
  \includegraphics[width=0.4\textwidth]{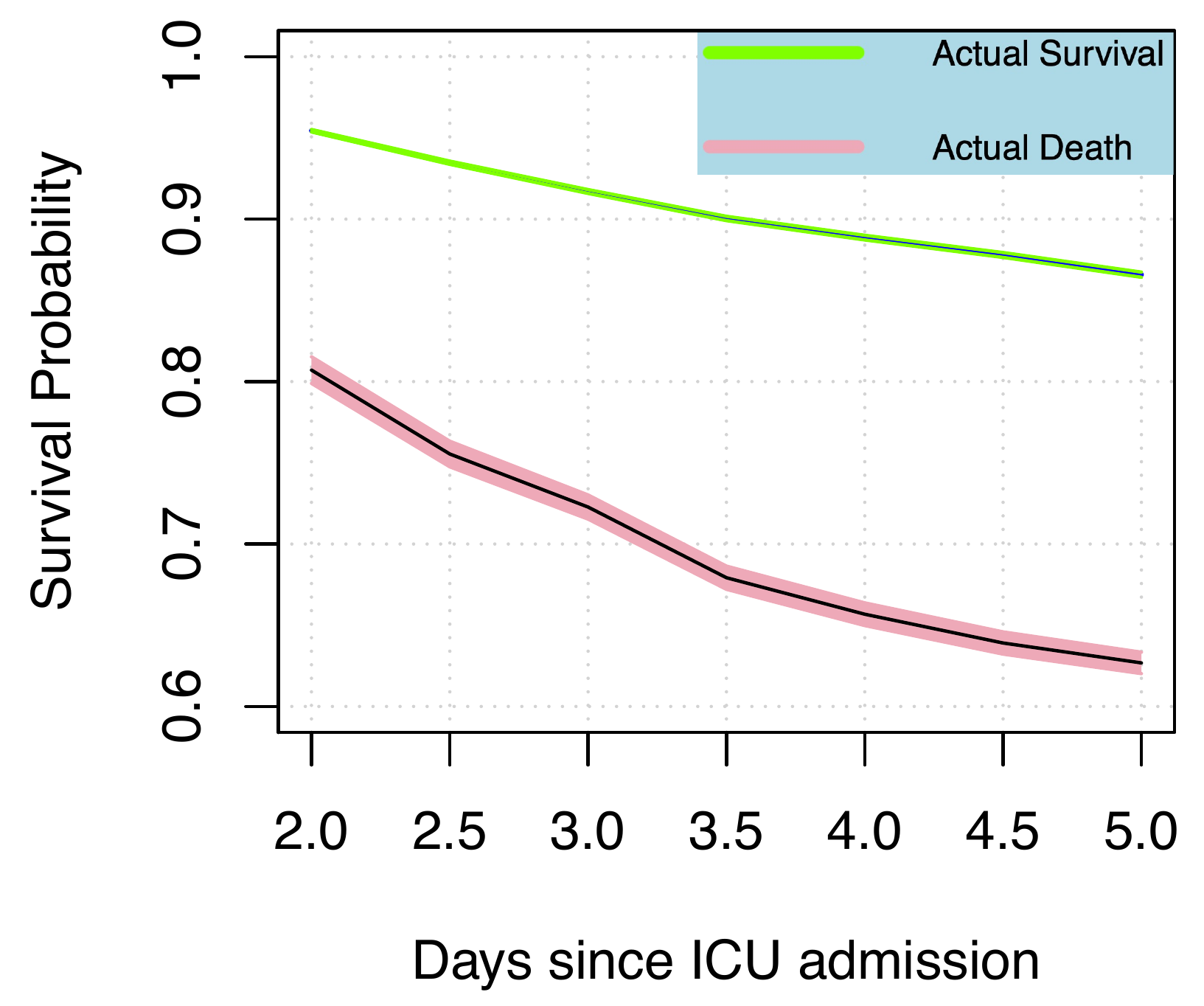} 
  \caption{ The survival curve is plotted based on CHF-AR-HMM's survival probability. The green area is the 95\% confidence interval for actual survival patients. The pink area is the 95\% confidence interval for actual death patients. The black lines are the mean survival probability of both actual survival and death patients.}
  \label{surv}
\end{figure}

The performance on the early mortality prediction for respiratory failure patients from Day 2 to Day 5 after ICU admission has been shown in Table \ref{output_table}. All the performance outcomes from cross-validation have been tested by paired one-tailed t-test. For all target days, performance of CHF-AR-HMM has the statistically significant improvement with all p-value less than 0.02. On Day 2, the CHF-AR-HMM has significantly improved the performance on AUCPR by 2\% (p-value $<$ 0.02) comparing with exponential parametric survival regression, which is the best performed baseline method. On Day 5, comparing with exponential parametric survival regression, the CHF-AR-HMM has significantly improved the performance on AUCPR by 4\% (p-value $<$ 0.001). CHF-AR-HMM has significantly 1\% (p-value $<$ 0.01) better than C-statistic than exponential parametric survival regression on Day 2 mortality prediction. On Day 5, CHF-AR-HMM has 2\% (p-value  $<$ 0.001) significant improvement comparing to exponential parametric survival regression. For AUROC, the CHF-AR-HMM has 2\% (p-value  $<$ 0.02) significant improvement on Day 2 comparing to exponential parametric survival regression. On Day 5, the AUROC from CHF-AR-HMM is 2\% ($p$-value  $<$ 0.001) higher than that from exponential parametric survival regression. For survival analysis, the CHF-AR-HMM estimated survival probability generated from $\eta$ is displayed in Figure \ref{surv} along the target days. The estimated survival probability curve for actual death patients is lower than that from the actual survival patients.

\section{CONCLUSIONS}
\label{seciv}
In this study, CHF-AR-HMM is used for the prediction of early mortality for the ICU patients with respiratory failure.
For all target days, comparing with baseline methods, our model shows significantly better capability of classifying the death class. While the prediction capability of our method is also better for both classes with equal importance. Meanwhile, the CHF-AR-HMM has significantly higher concordance comparing with other baseline methods. The survival curves are the distinguishable for actual death and survival patients as shown Figure \ref{surv}. In the future work, we seek to enhance the positive class prediction capability by reducing the data overlapping with further feature engineering.

\bibliographystyle{IEEEtran}
\bibliography{myref.bib}

\end{document}